\documentclass[runningheads]{llncs}

\usepackage{graphicx}
\usepackage{hyperref}
\hypersetup{colorlinks=true,allcolors=.,urlcolor=blue}
\usepackage{booktabs}
\usepackage{multicol}
\usepackage{multirow}
\usepackage{todonotes}
\usepackage{csquotes}
\usepackage[english]{babel}

\begin{document}
\title{Open Source Handwritten Text Recognition\\
on Medieval Manuscripts using Mixed Models\\
and Document-Specific Finetuning}
\titlerunning{Open Source Handwritten Text Recognition on Medieval Manuscripts}
%
\author{
Christian Reul\inst{1}
\and Stefan Tomasek\inst{1}
\and Florian Langhanki\inst{1}
\and Uwe Springmann\inst{2}
}
\authorrunning{C. Reul et al.}
%
\institute{University of Würzburg, Germany\\
\email{christian.reul@uni-wuerzburg.de}\\
\email{stefan.tomasek@germanistik.uni-wuerzburg.de}\\
\email{florian.langhanki@uni-wuerzburg.de}\\
\and
CIS, LMU Munich, Germany\\
\email{springmann@cis.uni-muenchen.de}
}
\maketitle              

\begin{abstract}

This paper deals with the task of practical and open source Handwritten Text Recognition (HTR) on German medieval manuscripts.
We report on our efforts to construct mixed recognition models which can be applied out-of-the-box without any further document-specific training but also serve as a starting point for finetuning by training a new model on a few pages of transcribed text (ground truth). To train the mixed models we collected a corpus of 35 manuscripts and ca.\ 12.5k text lines for two widely used handwriting styles, Gothic and Bastarda cursives. Evaluating the mixed models out-of-the-box on four unseen manuscripts resulted in an average Character Error Rate (CER) of 6.22\%. After training on 2, 4 and eventually 32 pages the CER dropped to 3.27\%, 2.58\%, and 1.65\%, respectively.
While the in-domain recognition and training of models (Bastarda model to Bastarda material, Gothic to Gothic) unsurprisingly yielded the best results, finetuning
out-of-domain models to unseen scripts was still shown to be superior to training from scratch. 

Our new mixed models have been made openly available to the community.

\end{abstract}
\section{Introduction}
\label{sec:introduction}

Efficient methods of Automatic Text Recognition (ATR) of images of either printed (OCR) or handwritten (HTR) material rely on the availability of pretrained recognition models that have been trained on a wide variety of various glyphs (the specific incarnations of an alphabet of characters on paper) that one wants to recognize.
Appropriately trained \textit{mixed models} have shown to result in Character Error Rates (CERs) below 2\% even for historical printed material previously unaccessible to automated recognition \cite{reul2021mixed}.
Automatically transcribed text resulting from the out-of-the-box application of available mixed models will already enable use cases such as searching and can enable corpus-analytic studies if large text collections are available.

For other downstream tasks such as determining that certain words are not contained in a text or preparing a critical edition, much lower error rates have to be achieved.
Before embarking on a time-consuming manual correction project, automatic methods can again come to the rescue to a certain extent.
Following the lead of Breuel et al. \cite{breuel2013lstm} it was shown that printing-specific neural-network-based models can be trained that deliver better results than mixed models \cite{springmannluedeling2017} which in turn have been shown to provide an exellent starting point for the training procedure \cite{reul2018combining}.

In the end, even for the result of these fine-tuned models it may still be necessary to manually correct the outcome.
However, the best way to achieve one's goal of a certain acceptable residual error rate (no printed text can ever be 100\% error-free) with minimal human effort is to combine out-of-the-box recognition with a so-called Iterative Training Approach (ITA) which has already been shown to be highly effective when dealing with early printed books \cite{reul2019ocr4all}.
The idea is to continously retrain specialized models on manually transcribed lines (Ground Truth, GT), apply them to new data, and use the ever-increasing accuracy to keep the error rates and therefore the required correction effort as low as possible at all times.

While the above method has previously been shown to work well with printed material, we here report on experiments with medieval German manuscripts.
Because handwritten material is much less regular than printed text, resulting from variation of glyphs among different writers and even the same writer, we expect considerably higher error rates.
Just how low an error rate can be achieved by the application of a pretrained mixed model that gets finetuned to a specific document with moderate effort (a few pages of GT) is the topic of the current paper.

A second goal is to explore what effort it takes to train a finetuned model based on a mixed model.
As we make our models openly available\footnote{\url{https://github.com/Calamari-OCR/calamari_models_experimental}}, the question is: What increase in recognition quality can be achieved by continuously training a finetuned model on an ever larger amount of generated GT?

The third goal explores the possibility of adapting a mixed model out of the domain of the training material it is based on.
Would it still be possible to finetune a mixed model with a few pages of GT to an out-of-domain manuscript successfully?

The remainder of the paper is structured as follows:
After an overview of related work in Section \ref{sec:related-work} we introduce the data required for training and evaluation in Section \ref{sec:data} and explain the methodology of our experiments in Section \ref{sec:methods}.
The experiments are described in Section \ref{sec:experiments} and discussed in Section \ref{sec:discussion} before Section \ref{sec:conclusion} concludes the paper.

\section{Related Work}
\label{sec:related-work}

For a thorough literature review of HTR methods we refer to a recent survey by Memon et al. \cite{memon2020handwritten} and to the description of a set of benchmarks by Sanchez et al. \cite{sanchez2019set}.

The Transkribus platform \cite{transkribus} offers a comprehensive selection of publicly available models\footnote{\url{https://readcoop.eu/transkribus/public-models}} for a variety of languages and different epochs. However, of the 85 models available at the time of writing, only 14 can be used with the open source\footnote{\url{https://github.com/jpuigcerver/PyLaia}} PyLaia engine \cite{mocholi2018development} while the vast majority (71 models) are only available for the proprietary HTR+ engine \cite{michael2018htr} which can exclusively be used via Transkribus. The vast majority of the public models were trained using texts from the 16\textsuperscript{th} century or later.

Hawk et al. (see \cite{hawk2019modelling}) reported on experiments with mixed models dealing with Caroline minuscle scripts using the OCRopus OCR engine\footnote{\url{https://github.com/ocropus/ocropy}}. 
Several experiments with varying numbers of different manuscripts in the training set showed that when applying a mixed model to a manuscript which has not been part of the training pool, models trained on a wider variety of manuscripts perform better.
However, when a model is applied to material that the model has already seen during training (same manuscript but different pages), the trend is mostly the reverse.

In \cite{stoekl2021biblia} Stökl Ben Ezra et al. present an open annotated dataset\footnote{\url{https://zenodo.org/record/5167263}} and pretrained script-specific as well as a mixed model for recognition and page segmentation on Medieval Hebrew manuscripts.

Hodel et al. \cite{hodel2021general} deal with mixed models for German \textit{Kurrent} script. They present an open test set\footnote{\url{https://zenodo.org/record/4746342}} comprising 2,426 lines collected from minutes of the meetings of the Swiss Federal Council between 1848 and 1903. Evaluating three HTR+ and one PyLaia Kurrent model on the test set resulted in median CER values from 2.76\% to 13.30\%.

\section{Data Sets}
\label{sec:data}

To perform our experiments we first had to collect and produce data, both as page images of medieval manuscripts as well as the corresponding GT in the form of diplomatic\footnote{\url{https://en.wikipedia.org/wiki/Diplomatics\#Diplomatic_editions_and_transcription}} (i.e.\ faithful to the written image) transcriptions.
The training and evaluation sets we compiled are described in this section.

To ensure maximum flexibility and connectivity we always collected the original color images and used the PAGE XML format \cite{pletschacher2010page} to store any further information like region and line coordinates, transcriptions, etc.

Our training corpus was collected and produced from various sources and within several projects as summarized in Table~\ref{tab:data_train}: The \textit{Kindheit Jesu (Childhood of Jesus)} editorial project at University of Würzburg, the \textit{Parzival} (\textit{Percival})\footnote{\url{https://www.parzival.unibe.ch/englishpresentation.html}} digital editorial project at University of Bern, some theological manuscripts transcribed during the \textit{Faithful Transcriptions} transcribathon\footnote{\url{https://lab.sbb.berlin/events/faithful-transcriptions-2/?lang=en}} \cite{Eichenberger2021Faithful}, several manu\-scripts containing the \textit{Marienleben (Life of Mary)} and some medieval medical tracts.

These manuscripts have been chosen to cover both a span of several centuries (13\textsuperscript{th} to 16\textsuperscript{th}) of origin as well as a certain variety of writing styles typical for this time: Gothic and Bastarda cursive.\footnote{\url{https://www.adfontes.uzh.ch/tutorium/schriften-lesen/schriftgeschichte/bastarda-und-gotische-kursive}}

Manuscript pages from the first two projects were manually transcribed while for the last two projects the segmentation of the images and the subsequent transcriptions was done via the open source OCR4all\footnote{\url{https://github.com/ocr4all}} framework.
All pre-existing transcriptions had to be adapted to our transcription guidelines to have a uniform representation of glyphs to characters.

\begin{figure}[!htb]
    \centering
    \includegraphics[width=\linewidth]{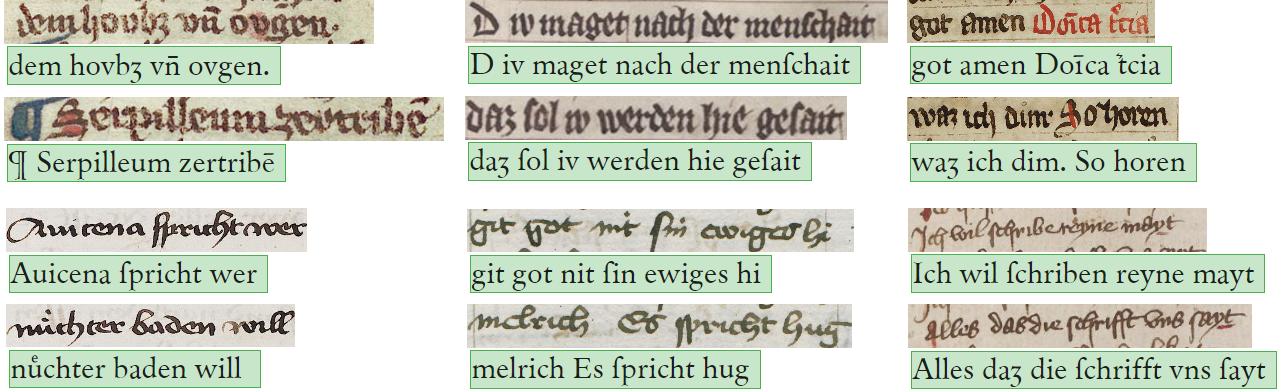}
    \caption{Two line images for each of six representative manuscripts and their corresponding transcription for the training corpus (top: Gothic, bottom: Bastarda).}
    \label{fig:data_train}
\end{figure}

In total, the training stock consists of 35 manuscripts with ca.\ 12,5k lines which can be broken down into close to 8,5k lines of Gothic and ca.\ 4k lines of Bastarda cursives.
Figure~\ref{fig:data_train} shows some representative example lines and their corresponding transcriptions.

\begin{table}[tb]
\centering
\caption{Training data used for our experiments.
Apart from the \textit{subcorpus} we list the number of \textit{manuscripts} as well as the respective number of \textit{pages} and \textit{lines} (all and selected for training).}
\label{tab:data_train}
\begin{tabular}{l|r|rr|rr|rrrr}
\toprule

\multirow{2}{*}{\textbf{subcorpus}} & \multirow{2}{*}{\textbf{\# works}} & \multicolumn{2}{c|}{\textbf{all}} & \multicolumn{2}{c|}{\textbf{selected}} & \multicolumn{4}{c}{\textbf{centuries}} \\
 & & \textbf{\# pages} & \textbf{\# lines} & \textbf{\# pages} & \textbf{\# lines} & \textbf{13} & \textbf{14} & \textbf{15} & \textbf{16} \\

\midrule

Kindheit Jesu           &  5 & 128 & 6,244 & 18 &   576 & 3 & 2 &  & \\
Parzival                &  6 &  36 & 1,685 & 19 &   771 &  &  & 6 & \\
Faithful Transcriptions & 12 &  73 & 2,483 & 44 & 1,381 &  & 2 & 9 & 1 \\
Marienleben             &  6 &  28 & 1,232 & 19 &   791 &  & 4 & 2 & \\
Medical Tracts          &  6 &  26 &   891 & 21 &   699 & 1 & 1 & 4 & \\

\midrule

Sum & 35 & 291 & 12,535 & 121 & 4,218 & 4 & 9 & 21 & 1 \\

\bottomrule
\end{tabular}
\end{table}

\begin{figure}[!htb]
    \centering
    \includegraphics[width=\linewidth]{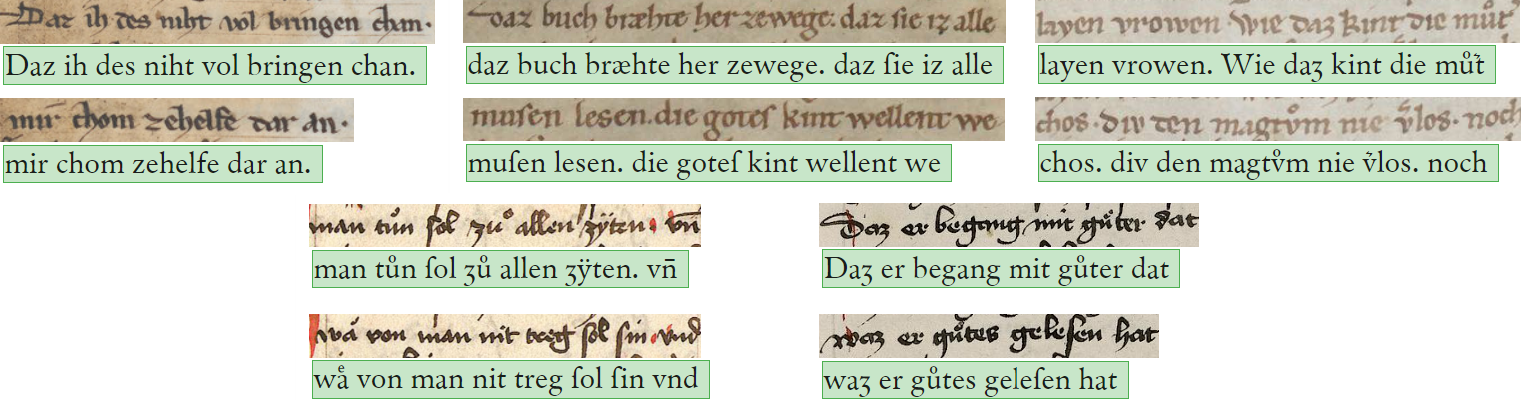}
    \caption{Two line images and the corresponding transcription for the five manuscripts used for evaluation. Top:  Handschrift-B, Wernher-Krakau, Wernher-Wien; bottom: Gast-1, Gast-2}
    \label{fig:data_eval}
\end{figure}

To evaluate our models we collected 212 pages comprising close to 9k lines from five additional manuscripts, three written in Gothic and two in Bastarda cursives (cf.\ Figure~\ref{fig:data_eval} for some example lines):
One manuscript containing the \textit{Kindheit Jesu} (Handschrift-B), two additional manuscripts about the life of Mary (\textit{Driu liet von der maget}) written by Brother Wernher (Wernher-Krakau and Wernher-Wien), and as Bastarda examples we chose two manuscripts of the moral doctrine \textit{Der Welsche Gast} written by Thomasin von Zirclaere\footnote{\url{https://digi.ub.uni-heidelberg.de/wgd}} (Gast-1 and Gast-2) from the digital editorial project of the University of Heidelberg.
Wernher-Wien was written by a hand already present in the training data and added for comparison.

For our experiments we randomly selected 32 pages as our maximum training set and then repeatedly cut it in half to obtain further sets comprising 16, 8, 4, and 2 pages.
The rest of the data we used as a fixed evaluation set.
Table~\ref{tab:data_eval} lists the details.

\begin{table}[tb]
\setlength{\tabcolsep}{0.5em}
\centering
\caption{Number of available \textit{pages} and \textit{lines} for the evaluation manuscripts, devided into a fixed evaluation set (\textit{Eval}) and five training batches for the ITA (\textit{Train}).}
\label{tab:data_eval}
\begin{tabular}{lc|rr|rrrrr}
\toprule

\multirow{3}{*}{\textbf{Manuscript}} & \multirow{3}{*}{\textbf{Date}} & \multicolumn{2}{c|}{\textbf{Eval}} & \multicolumn{5}{c}{\textbf{Train}} \\
 & & \textbf{\# pages} & \textbf{\# lines} & \multicolumn{5}{c}{\textbf{\# lines for \# pages}} \\
 & & & & 2 & 4 & 8 & 16 & 32\\

\midrule

Handschrift-B   & 1250-1275 &  8 &  592 & 152 & 304 & 607 & 1.215 & 2.430 \\
Wernher-Krakau  & 1200-1225 & 18 &  363 &  49 &  96 & 197 &   349 &   687 \\
Wernher-Wien    & 1250-1275 & 10 &  241 &  49 &  98 & 196 &   391 &   752 \\

\midrule

Gast-1 & 1450-1475 & 8 &  546 & 141 & 261 & 521 & 1.070 & 2.160 \\
Gast-2 & ca.\ 1300 & 8 &  222 &  58 & 114 & 217 &   442 &   881 \\

\bottomrule
\end{tabular}
\end{table}

\section{Methods}
\label{sec:methods}

Training mixed models which can be applied out-of-the-box as well as serve as a starting point for document-specific training is a challenging task: On the one hand, the models need to be very robust and generalize well on as many documents as possible. On the other hand, the models should still be quite specifically geared towards a certain type of material in order to produce the best possible results.

\begin{samepage}
To deal with this challenge, we devised the following pipeline where each training starts from the model resulting from the previous step:

\begin{enumerate}
    \item A strong mixed model for printed types based on \cite{reul2021mixed} serves as a foundation for all training processes.
    Despite the apparent difference to manuscript material we expect this to provide a better starting point compared to starting from scratch, i.e.\ a random parameter allocation.
    \item Both handwriting styles (Gothic and Bastarda) are trained together to show the model as much data as possible in order to learn general features, adapt to noise, etc. The result is a combined mixed model covering both styles.
    \item Finally, the existing models are refined by exclusively training on Gothic script or Bastarda data, respectively. This results in a mixed model for each handwriting style.
\end{enumerate}
\end{samepage}

An important factor when dealing with ATR or deep learning in general is the selection and optimization of hyperparameters which define the structure of the network.
We refrain from evaluating a wide variety of configurations, but stick to the following networks which have been shown to be very useful in the past and are predefined in the open source\footnote{\url{https://github.com/Calamari-OCR/calamari}} Calamari ATR engine \cite{wick2018calamari} which we used for our experiments:

\begin{itemize}
    \item \textit{def}: The original and comparatively shallow Calamari default network structure consisting of two convolutional neural networks (CNNs) (40 and 60 3x3 filters respectively), each followed by a 2x2 max pooling layer, and a concluding LSTM network layer with 200 cells using dropout (0.5)\footnote{In Calamari short notation:\newline
    conv=40:3x3,pool=2x2,conv=60:3x3,pool=2x2,lstm=200,dropout=0.5}.
    \item \textit{htr+}: An adaptation of the standard network structure of the Transkribus platform with more complex variations regarding filter sizes, strides, etc. (see  \cite{michael2018htr} for details).
    \item \textit{deep3}: An alternative deeper network structure which extends the default network by another convolutional layer and two additional LSTMs.
    This network yielded good results during our previous experiments with mixed models for printed material and is becoming the new Calamari default\footnote{In Calamari short notation:\newline
    conv=40:3x3,pool=2x2,conv=60:3x3,pool=2x2,conv=120:3x3,pool=2x2,\newline
    lstm=200,lstm=200,lstm=200,dropout=0.5}.
\end{itemize}

Analogously to the method described in \cite{reul2021mixed} we apply a two-stage training approach to reduce the influence of a few overrepresented manuscripts while utilizing all of the available training data:
For each individual step in the training workflow we first train on all available data, to show the model as much material as possible.
Then, during the so-called refinement stage, we use the resulting model from stage one as a starting point and run another full training process while only using selected pages for each manuscript.
The selected pages have been determined beforehand by simply drawing single pages at random until a predetermined number of lines is surpassed or all pages have been drawn for a single manuscript.
Since almost all works comprised more than or at least close to 150 lines of GT we chose this number as our cutoff value. In this way a balanced training corpus is constructed which gives equal weight to all manuscripts.

Second, we varied the input data by using different preprocessing results, i.e.\ mainly different binarization techniques, which can also be regarded as a form of data augmentation. We used two methods from the \textit{ocrd-olena} package\footnote{\url{https://github.com/OCR-D/ocrd_olena}} (Wolf, Sauvola MS Split), the binary and normalized grayscale output produced by OCRopus' \textit{ocropus-nlbin} script, as well as the SBB binarization technique\footnote{\url{https://github.com/qurator-spk/sbb_binarization}}.

As mentioned in the introduction, the most cost-effective way to get to a low-error output consists in avoiding manual correction work as much as possible by building better recognition models and thus replacing human effort by a higher but much cheaper computational load.
This issue of getting to a reasonable recognition accuracy as fast as possible is addressed by the so-called \textit{Iterative Training Approach} \cite{reul2019ocr4all} which consists of the following steps:

\begin{enumerate}
	\item Transcribe a small number of lines from scratch or correct the output of a suitable mixed model, if available
	\item Train a document-specific model using all available GT (including the GT from earlier iteratons)
	\item Apply the model to further lines which have not yet been transcribed
	\item Correct the output
	\item Repeat steps 2-4
\end{enumerate}

In this paper the ITA is utilized two times:
First, when transcribing manuscripts to produce initial GT for the evaluation set. 
Second, we simulate an ITA during the second experiment by iteratively doubling the training pages.

\section{Experiments}
\label{sec:experiments}

For our experiments we used Calamari version 2.1.
All training runs followed the cross-fold training methodology of \cite{reul2018voting} producing a voting ensemble of five individual voters and combining their respective outputs via the confidence values of each individual recognized character.
Apart from better results this also reduces the variance among the results considerably.
In addition, we used the same random seed for the experiments to standardize all processes involving randomness (data shuffling, augmentation etc.).

To determine the end of each training process we utilized the default early stopping criterion which evaluates the current model against the held out validation data after each epoch.
The training stops if the validation CER did not improve for five consecutive times but at the latest after 100 epochs.
The number of samples after which an evaluation was performed was set according to the size of the respective training set but always to at least 1,000 steps.
Regarding further (hyper-) parameters we stuck to the Calamari defaults and well established best practices.
Most notably this includes augmenting each sample five times by using image degradation transformations for each written line as well as utilizing a general weight decay of $10^{-5}$ for all layers and an EMA weight decay of 0.99.

The model recognition output is evaluated against the GT by the Levenshtein edit distance which measures the CER. 
As for preprocessing, we exclusively use the sbb binarization output for recognition and training in our experiments since it has been proven to provide good results for a variety of image conditions.

\subsection{Determining the best Starting Model}

We first performed only a few selected experiments to determine the best general approach regarding the network structure and the degree of generalization of the mixed models.

\begin{enumerate}
    \item Start off with a small amount of GT to obtain a first book-specific model which hopefully already considerably outperforms the initial mixed model.
    \item Build a strong specialized model by applying the ITA procedure of iteratively adding more training data to the training pool. This model's output may already be sufficient for many use-cases
    and allows to efficiently weed out the remaining errors manually.
\end{enumerate}

All training processes are performed for each of the three network structures introduced above (default, htr+, and deep3).
To examine the influence of the different handwriting styles we compare the output of the combined, Bastarda and Gothic models trained according to Sect. \ref{sec:methods}.
Despite these variations, all trainings follow the two-stage training procedure: 1) train on all available pages, 2) use the output from 1) and refine it on the more balanced set of selected pages.

We evaluate the resulting models on two selected works, i.e.\  Wernher-Krakau (Gothic script) and Gast-1 (Bastarda).
For each of the works we define a subset of evaluation pages which remain unchanged for all upcoming experiments to ensure comparability.
Table~\ref{tab:exp_networks} sums up the results.

\begin{table}[tb]
\centering
\caption{The upper part of the table lists the CERs (in \%) for the two manuscripts Wernher-Krakau (Gothic) and Gast-1 (Bastarda) recognized with the three pretrained mixed models (combined, Bastarda, and Gothic). The CERs are averaged results for out-of-the-box and after document-specific training on 4 and 16 pages and given for three different \textit{network} structures (\textit{default}, \textit{htr+}, and \textit{deep3}).
Finally, the CERs are averaged again for each network (last column) and each pretrained model (last row).
The respective best values are marked in bold.
The lower part of the table lists the individual CERs after training from scratch (no PT = no pretraining) and with pretrained models (out-of-the-box, after training on 4 and 16 pages) for the deep3 network.}
\label{tab:exp_networks}
\begin{tabular}{cccccccccc}
\toprule

 & \multicolumn{4}{c}{\textbf{Wernher-Krakau}} & \multicolumn{4}{c}{\textbf{Gast-1}} & \multirow{2}{*}{Avg.} \\

Network &  & combined & Bastarda & Gothic &  & combined & Bastarda & Gothic  \\

\midrule

default &  &  4.62 &  9.16 &  \textbf{4.44} &  & 6.23 &  \textbf{5.85} & 16.51 & 7.80 \\

htr+    &  & \textbf{2.98} &  6.79 &  3.23 & &  4.42 &  \textbf{4.24} & 12.25 & 5.65 \\

deep3   &  &  3.15 &   6.88 &  \textbf{3.05} &  &  \textbf{3.82} &  3.97 &  10.05 & \textbf{5.15} \\

\midrule

Avg. &  & 3.58 &   7.61 &  \textbf{3.57} &  &  4.82 &  \textbf{4.69} &  12.93 & - \\

\midrule
\midrule

\multicolumn{10}{l}{Detailed results for the deep3 network}\\

Network & no PT & combined & Bastarda & Gothic & no PT & combined & Bastarda & Gothic  \\
\midrule
ootb &    - & 6.53 &16.56 & 6.21 &    - & 6.92 & 7.24 &24.88 &  \\
   4 & 5.27 & 1.62 & 2.51 & 1.59 & 8.33 & 2.71 & 2.80 & 3.28 &  \\
  16 & 2.29 & 1.31 & 1.56 & 1.35 & 2.89 & 1.82 & 1.88 & 1.99 &  \\

\bottomrule
\end{tabular}
\end{table}

As expected, the two deeper networks perform considerably better than the shallow one with deep3 achieving the lowest average CER (5.15\% compared to 5.65\%/7.80\% for def/htr+).
Applying Bastarda models to Bastarda material and Gothic to Gothic achieves the best results, closely followed by the combined model.
Based on these outcomes we will perform the upcoming experiments using deep3 as our network structure and always apply the most-suitable (like-for-like) model.

\subsection{Iterative Document-Specific Training}

After determining the best starting model we want to take a closer look at the behaviour of the models produced during the ITA.
This experiment is carried out on all five available evaluation manuscripts.
To simulate the ITA we define fixed training pages for each iteration of the training process.
For practical reasons we always stick to full pages independent of their number of lines or tokens.
We start out with a very managable amount of GT consisting of only two pages and always double that number during the following iterations (2, 4, 8, 16, and 32 pages) where each set of pages completely subsumes the previous set.
Note that each iteration starts from the initial mixed model and not from the model produced during the previous iteration.
This is important, since we expect the learned \textit{knowledge} of the mixed model to gradually diminish during more specific training. By beginning the training from the initial model each time we hope to counteract this forgetting effect.
Each book-specific model as well as the initial mixed model is applied to a fixed evaluation set, i.e.\ all pages not used for training.
Table~\ref{tab:exp_ita} sums up the results.

\begin{table}[tb]
\setlength{\tabcolsep}{0.3em}
\centering
\caption{Each row lists the CERs and improvement rates (\%) obtained by training on the given number of pages (\textit{\# pages}) with row \textit{0} showing the out-of-the-box results.
Each training was performed by starting from scratch (column \textit{FS}) and by utilizing pretrained model as a starting point (\textit{PT}).
Column \textit{Impr.} shows the improvement of PT over FS and of PT over the previous iteration.
Wernher-Wien is special since it was produced by a hand already present in the training data and thus represents a true in-domain application of our pretrained model. It is therefore not included in the average values given in the lower last column \textit{Avg. All}.
}
\label{tab:exp_ita}
\begin{tabular}{c|ccc|ccc|ccc|ccc}
\toprule

 & \multicolumn{3}{c}{\textbf{Wernher-Krakau}} & \multicolumn{3}{c}{\textbf{Handschrift B}} & \multicolumn{3}{c}{\textbf{Avg. Gothic}} & \multicolumn{3}{c}{\textbf{Wernher-Wien}} \\
\# pages & FS & PT & Impr. & FS & PT & Impr. & FS & PT & Impr. & FS & PT & Impr. \\

\midrule
 
 0 &     - & 6.21 &    - &     - & 4.90 &    - &     - & 5.55 &    - &     - & 2.99 &   - \\
 \midrule
 2 & 13.67 & 1.95 & 86/69 & 10.73 & 2.61 & 76/47 & 12.20 & 2.28 & 81/59 & 22.42 & 2.39 & 89/20 \\
\midrule
 4 &  5.27 & 1.59 & 70/19 &  7.68 & 2.30 & 70/12 &  6.48 & 1.95 & 70/15 &  7.95 & 2.11 & 74/12 \\
\midrule
 8 &  2.57 & 1.45 & 44/9  &  4.07 & 1.89 & 54/18 &  3.32 & 1.67 & 50/14  &  4.24 & 1.99 & 53/6 \\
\midrule
16 &  2.29 & 1.35 & 41/7  &  3.81 & 1.64 & 57/13 &  3.05 & 1.50 & 51/11  &  3.10 & 1.84 & 41/8 \\
\midrule
32 &  1.56 & 1.31 & 16/3  &  3.30 & 1.38 & 58/16 &  2.43 & 1.35 & 45/10 &  2.33 & 1.57 & 33/15 \\
\midrule

\midrule
 
 & \multicolumn{3}{c}{\textbf{Gast-1}} & \multicolumn{3}{c}{\textbf{Gast-2}} & \multicolumn{3}{c}{\textbf{Avg. Bastarda}} & \multicolumn{3}{c}{\textbf{Avg. All}} \\
\# pages & FS & PT & Impr. & FS & PT & Impr. & FS & PT & Impr. & FS & PT & Impr. \\

\midrule

 0 &     - & 7.24 &    -  &     - & 6.54 &    - &      - & 6.89 &     - &     - & 6.22 &   - \\
 \midrule
 2 & 16.69 & 3.50 & 79/52 & 43.39 & 5.00 & 89/24 & 30.04 & 4.25 & 86/38 & 21.12 & 3.27 & 85/48 \\
\midrule
 4 &  8.33 & 2.80 & 66/20 & 21.68 & 3.61 & 83/28 & 15.01 & 3.21 & 79/25 & 10.74 & 2.58 & 76/21 \\
\midrule
 8 &  4.65 & 2.26 & 51/19 & 11.98 & 3.09 & 74/14 &  8.32 & 2.68 & 68/17 &  5.82 & 2.17 & 63/16 \\
\midrule
16 &  2.89 & 1.88 & 35/17 &  6.12 & 2.87 & 53/7  &  4.51 & 2.38 & 47/11 &  3.78 & 1.94 & 49/11 \\
\midrule
32 &  2.26 & 1.64 & 27/13 &  3.73 & 2.25 & 40/22 &  3.00 & 1.95 & 35/18 &  2.71 & 1.65 & 39/15 \\

\bottomrule
\end{tabular}
\end{table}

\section{Discussion}
\label{sec:discussion}

Inspecting column \textit{Avg. All} of Table \ref{tab:exp_ita} shows that
applying the pretrained mixed models out-of-the-box achieves an average CER of 6.22\% which then quickly improves during finetuning following the ITA.
Just two pages of GT are enough to achieve a CER of 3.27\% improving the out-of-the-box output by 48\% on average.
As expected, further iterations lead to further improvements, resulting in an average CER of 1.65\% when utilizing 32 pages of GT.
Using a pretrained model as a starting point for the document-specific trainings leads to significantly lower error rates compared to starting from scratch, with the improvement factors diminishing with more pages of GT being added, ranging from 85\% (2 pages) to 39\% (32 pages).
Finally, the CERs for the manuscripts written using Gothic handwriting (excluding Wernher-Wien) are considerably lower than the ones of the Bastarda manuscripts, both for the out-of-the-box recognition (5.56\% for Gothic fonts, 6.89\% for Bastarda ones) as well as for document-specific training (on average the CERs for Gothic is about 35\% lower compared to Bastarda).

Overall, these results are very promising.
The CER achieved by applying the pretrained mixed models out-of-the-box (on average 6.22\% for both scripts combined) is already good enough for some downstream tasks like (error-tolerant) full-text-search and certainly allows for a much quicker GT production compared to transcribing from scratch.
The subsequent document-specific training utilizing the mixed models as a starting point quickly led to vast improvements even when using just a few pages of GT.
Two pages were enough to instantly improve the out-of-the-box output considerably (48\% on average), resulting in a CER of 3.27\%.
These small amounts of GT can easily be produced by a single researcher, especially when starting from an already quite low-error recognition output of a mixed model or a document-specific model trained during an earlier iteration.

A thorough document-specific training using 32 pages of GT yielded an excellent average CER of 1.65\% with four out of the five evaluation manuscript reaching CERs of considerably below 2\%. The only exception is Gast-2 (CER of 2.25\%) which was somewhat expected since it represents the most challenging work and each page only comprises less than thirty rather short lines.
For comparison, this is about 2.5 times less than the other bastard manuscript Gast-1.
The high quality of HTR output not only enables a wide variety of possibilities for subsequent use of the generated texts but can also serve for further very efficient GT production, as the ITA does not have to stop here.
In fact, increasing the number of pages used for training from 16 to 32 still yielded a very notable improvement factor of 15\%, indicating further room for improvement.
However, the overall return on investment, i.e.\ the gains in CER in relation to the necessary amount of human effort, is diminishing when more and more material-specific GT is produced and trained.

Since the ITA can eventually be continued forever, it is up to the users to choose the strategy best suited to their material, use-case, and quality requirements.
For example, when the goal is to transcribe an entire manuscript to prepare a critical editions, which requires (basically) error-free text, it naturally makes sense to continue material-specific training until the transcription is finished.
On the contrary, when the goal is to reach a certain target CER required to perform specific downstream tasks the ITA should of course only be used until this target is reached to then process the remaining pages fully automatically.

Either way, the results show that utilizing a pretrained model as a starting point for the ITA is almost mandatory, as the effect compared to starting from scratch is immense. When using just two pages of training material the CERs improves by 85\% on average when incorporating a pretrained model. As expected, this effect decreases with a rising number of training pages. Yet, even when using a considerable amount of GT, i.e.\ 32 pages, the CER still improves by a very notable 39\%.

Most trainings can be completed within a couple of hours even when using a standard desktop PC without a GPU, making this approach highly feasible for the practicing humanist. This answers the first and second question from the introduction: Just training an available pretrained model on a few pages of GT that can be transcribed within a few hours, may provide recognized text with an error level in the low single digits. If the amount of training material is enlarged, the errors can further be diminished, albeit at a lower improvement rate.

This leaves the third question of what to do if only an out-of-domain pretrained model is available: Should one train from scratch on newly transcribed GT alone or would it still be better to start training with an existing out-of-domain model? A first indication that this might indeed be the case is apparent from comparing the results on Wernher-Wien whose Gothic hand was already present in the pretrained Gothic model (Table~\ref{tab:exp_ita}): While the out-of-the-box recognition is the best among all Gothic manuscripts, this advantage quickly vanishes with increasing document-specific training. 
This impression is reinforced by the out-of-domain experiments in Table \ref{tab:exp_networks} where, as expected, applying out-of-domain models (Bastarda to Gothic and vice versa) leads to considerably worse results.
Using out-of-domain models as a starting point for document-specific training still yields far superior results compared to starting from scratch, especially when not a lot of document-specific GT is available:
Training the out-of-domain Bastarda model (with an out-of-the-box CER of 16.56\%) resulted in vast improvements (2.51\%/1.56\%) even compared to training from scratch (5.27\%/2.29\%).
The same tendencies can be observed for the Bastarda manuscript Gast-1 with a Gothic out-of-the-box result of 24.88\% and trained results of 3.28\%/1.99\%, compared to training from scratch of 8.33\%/2.89\%.
The takeway from these observations is that users should not shy away from working with material for which no perfectly suited mixed model is available, but simply use the closest match and work from there.

Finally, we take a look at the change of the error distribution during the ITA  exemplified with the Wernher-Krakau manuscript (Table \ref{tab:exp_confusions}).
The biggest source of error in the out-of-the-box output are the dots at the end of each verse which are very small and often merge with the preceeding letter.
Further dominant errors are the confusion of a \textit{w} for \textit{vv} as well as the deletion of whitespaces and diacritical marks like the superscript \textit{e}, \textit{v}, and the abbreviation mark for \textit{er} (represented as a hook or a zig-zag sign above; marked as \textit{@} in the confusion table).
Document-specific training on 4 pages considerably improves the recognition of whitespaces and pushes almost all \textit{w}/\textit{vv} confusions and most diacritical errors beyond the ten most frequent errors, and close to 80\% of the dot-related errors have vanished. After training on 16 pages, dots and whitespace-related errors are responsible for the vast majority (close to 35\%) of remaining errors.

\begin{table}[tb]
\setlength{\tabcolsep}{3pt}
\centering
\caption{The ten most common confusions for 18 pages of the Wernher-Krakau manuscript showing the \textit{GT}, prediction (\textit{PRED}), the absolute number of occurrences of an error (\textit{CNT}) and its fraction of total errors in percent (\textit{\%}), resulting from the \textit{out-of-the-box} application of a mixed model and document-specific training using \textit{4/16 pages}.}
\label{tab:exp_confusions}
\begin{tabular}{ccrr|ccrr|ccrr}

\toprule

\multicolumn{4}{c|}{\textbf{out-of-the-box}} & \multicolumn{4}{c|}{\textbf{4 page training}} & \multicolumn{4}{c}{\textbf{16 page training}}\\

\textbf{GT} & \textbf{PRED} & \textbf{CNT}  & \textbf{\%} & 
\textbf{GT} & \textbf{PRED} & \textbf{CNT}  & \textbf{\%} &
\textbf{GT} & \textbf{PRED} & \textbf{CNT}  & \textbf{\%}\\

\midrule

.  &    & 121 & 13.2 & . &   & 25 & 10.7 & . &   & 24 & 12.1 \\
\textsuperscript{e} &    &  62 &  6.7 &   & . & 14 &  6.0 &   & ␣ & 17 &  8.6 \\
w  & vv &  44 &  9.6 & ␣ &   & 14 &  6.0 &   & . & 15 &  7.6 \\
␣  &    &  35 &  3.8 &   & ␣ & 12 &  5.2 & ␣ &   & 12 &  6.1 \\
\textsuperscript{v} &    &  33 &  3.6 & i &   &  9 &  3.9 & i &   & 10 &  5.1 \\
@ &    &  28 &  3.0 & z & l &  7 &  3.0 & in& m &  3 &  3.0 \\
z  & l  &  26 &  2.8 & \textsuperscript{e}&   &  6 &  2.6 & æ & a &  3 &  1.5 \\
d  & c  &  24 &  2.6 & t &   &  5 &  2.1 & n & u &  3 &  1.5 \\
u  & i  &  15 &  1.6 & i & u &  4 &  1.7 & \textsuperscript{v}&   &  3 &  1.5 \\
   & .  &  14 &  1.5 & n & u &  4 &  1.7 &   & \textsuperscript{e}&  3 &  1.5 \\



\bottomrule

\end{tabular}
\end{table}

\section{Conclusion and Future Work}
\label{sec:conclusion}

After creating a training corpus comprising 35 Medieval German manuscripts and close to 13k lines for two widely used German medieval handwritten styles, Gothic and Bastarda, we were able to train several highly performant mixed models.
Evaluation on four previously unseen manuscripts yielded very low error rates, both for the out-of-the-box application of mixed models (average CER below 6\%) as well as for document-specific training.
For the latter, the quality of the result strongly depended on the amount of training material used, ranging from average CERs of 3.28\% for just two pages of GT to 1.68\% for thoroughly trained models (32 pages).
A large share of this efficiency can be attributed to using the mixed models as a starting point for each individual material-specific training.
Pretraining showed to be highly effective with average improvement rates ranging from 86\% to 38\% depending of the number of pages used for training.
This not only held true for the in-domain application of models (Bastarda model to Bastarda material, Gothic to Gothic) but also when using out-of-domain models as a foundation for finetuning.

Better and more widely applicable pretrained models can be constructed the more GT is made openly available by individual researchers and groups. To foster this spirit of open collaboration we made our own pretrained models openly available.

Regarding future work, we want to utilize the intrinsic confidence values of the ATR engine.
For example, these confidence values could be used to identify individual lines the existing model struggled with the most and then transcribe these lines in a targeted way to maximize the training effect in an active learning-like approach. A first implementation for this, highlighting uncertain characters, has already been made available via OCR4all \cite{reul2019ocr4all}. In addition, aggregated confidence information could also provide an indicator for the current text quality and therefore serve as a stopping criterium for the ITA.

Current research strongly pushes towards recurrent-free approaches based on attention and transformer networks. Kang et al. \cite{kang2020pay} first introduced a non-recurrent architecture using multi-head self-attention layers not only on a visual but also on a textual level. Many refinements and combinations with existing approaches, e.g. \cite{diaz2021rethinking}, have followed since. While a broad application of these developments in the practical area is still hindered by the lack of stable and resource-efficient open source implementations, it seems clear that this is the direction into which ATR is heading in the future.

\subsection*{Acknowledgements}

The authors would like to thank our student research assistants Lisa Gugel, Kiara Hart, Ursula Heß, Annika Müller, and Anne Schmid for their extensive segmentation and transcription work as well as Maximilian Nöth and Maximilian Wehner for supporting the data preparation.

This work was partially funded by the German Research Foundation (DFG) under project no. 460665940.

\bibliographystyle{splncs04}
\bibliography{references}

\end{document}